\documentclass[10pt,twocolumn,letterpaper]{article}

\usepackage{iccv}
\usepackage{times}
\usepackage{epsfig}
\usepackage{graphicx}
\usepackage{amsmath}
\usepackage{amssymb}
\usepackage[accsupp]{axessibility}  %

\usepackage{ulem}

\newcommand{\uit}[1]{\uline{\textit{#1}}}
\usepackage{multirow}
\usepackage{booktabs}
\usepackage{color}
\usepackage[page]{appendix}

\definecolor{red}{rgb}{1,0,0}  
\definecolor{green}{RGB}{0, 133, 21}  
\definecolor{grey}{rgb}{0.5,0.5,0.5}

\newcommand{\diffguard}{\textsc{DiffGuard}\xspace}

\newcommand{\cifar}{\textsc{Cifar-10}\xspace}

\newcommand{\cifarh}{\textsc{Cifar-100}\xspace}

\newcommand{\imagenet}{\textsc{ImageNet}\xspace}

\newcommand{\tinyimagenet}{\textsc{TinyImageNet}\xspace}

\newcommand{\ltwo}{\ensuremath{\ell_2}\xspace}

\makeatletter
\def\blfootnote{\xdef\@thefnmark{}\@footnotetext}
\makeatother

\usepackage[breaklinks=true,bookmarks=false]{hyperref}

\iccvfinalcopy %

\def\iccvPaperID{2520} %

\ificcvfinal\fi

\begin{document}
\normalem %

\title{\diffguard: Semantic Mismatch-Guided Out-of-Distribution Detection \\using Pre-trained Diffusion Models}

\author{
    Ruiyuan Gao$^{\dag}$
    \quad
    Chenchen Zhao$^{\dag}$
    \quad
    Lanqing Hong$^{\ddag}$
    \quad
    Qiang Xu$^{\dag}$
    \\
    $^{\dag}$The Chinese University of Hong Kong
    \enspace
    $^{\ddag}$Huawei Noah's Ark Lab
    \\
    {\tt\small \{rygao, cczhao, qxu\}@cse.cuhk.edu.hk
    \quad
    honglanqing@huawei.com} 
}

\maketitle
\ificcvfinal\thispagestyle{empty}\fi

\begin{abstract}
Given a classifier, the inherent property of semantic Out-of-Distribution (OOD) samples is that their contents differ from all legal classes in terms of semantics, namely \emph{semantic mismatch}. There is a recent work that directly applies it to OOD detection, which employs a conditional Generative Adversarial Network (cGAN) to enlarge semantic mismatch in the image space. 
While achieving remarkable OOD detection performance on small datasets, it is not applicable to \imagenet-scale datasets due to the difficulty in training cGANs with both input images and labels as conditions. 

As diffusion models are much easier to train and amenable to various conditions compared to cGANs, in this work, we propose to directly use pre-trained diffusion models for semantic mismatch-guided OOD detection, named \diffguard. Specifically, given an OOD input image and the predicted label from the  classifier, we try to enlarge the semantic difference between the reconstructed OOD image under these conditions and the original input image. We also present several test-time techniques to further strengthen such differences. Experimental results show that \diffguard is effective on both \cifar and hard cases of the large-scale \imagenet, and it can be easily combined with existing OOD detection techniques to achieve state-of-the-art OOD detection results.
\blfootnote{Code: \url{https://github.com/cure-lab/DiffGuard}}

\end{abstract}

\section{Introduction}

The effectiveness of deep learning models is largely contingent on the independent and identically distributed (i.i.d.) data assumption, \ie, test sets follow the same distribution as training samples~\cite{krizhevsky2017imagenet}. However, in real-world scenarios, this assumption often does not hold true~\cite{drummond2006open}. Consequently, the task of out-of-distribution (OOD) detection is essential for practical applications, so that OOD samples can be rejected or taken special care of without harming the system's performance~\cite{hendrycks2017a}. 

For image classifiers, a primary objective of OOD detection is to identify samples having semantic shifts, whose contents differ from all legal classes in the training dataset~\cite{yang2021oodsurvey}. To differentiate such OOD samples and in-distribution (InD) ones, 
some existing solutions utilize information from the classifier itself, such as internal features~\cite{sun2022out}, logits~\cite{pmlr-v162-hendrycks22a}, or both~\cite{haoqi2022vim}. While simple, these solutions inevitably face a trade-off between the InD classification accuracy and the over-confidence of the trained classifier for OOD detection~\cite{nguyen2015deep}, especially on hard OOD inputs. Some other methods propose using an auxiliary module for OOD detection based on either reconstruction quality~\cite{denouden2018improving} or data density~\cite{pidhorskyi2018generative}. The auxiliary module does not affect the training process of the classifier, but these methods tend to have a low OOD detection capability.

To the best of our knowledge, MoodCat~\cite{yang2022out} is the only attempt that directly models the semantic mismatch of OOD samples for detection. Specifically, it employs a conditional Generative Adversarial Network (cGAN) to synthesize an image conditioned on the classifier's output label together with the input image. For InD samples with correct labels, the synthesis procedure tries to reconstruct the original input; while for OOD samples with semantically different labels, ideally the synthesis result is dramatically different from the input image, thereby enabling OOD detection. While inspiring, due to the difficulty in cGAN training with potentially conflicting conditions, MoodCat is not applicable to \imagenet-scale datasets.

Recently, diffusion models have surpassed GANs in terms of both training stability and generation quality. Moreover, they are amenable to various conditions during generation, including both label conditions~\cite{song2021scorebased,ho2021classifierfree} and image-wise conditions through DDIM inversion~\cite{song2021denoising}. 
With the above benefits, we propose a new semantic mismatch-guided OOD detection framework based on diffusion models, called \diffguard. 
Similar to~\cite{yang2022out}, \diffguard takes both the input image and the classifier's output label as conditions for image synthesis and detects OODs by measuring the similarity between the input image and its conditional synthesis result. 

However, it is non-trivial to apply diffusion models for semantic mismatch identification.  A critical problem with label guidance in diffusion models is the lack of consideration for the classifier-under-protection. This issue arises in both types of guidance in diffusion models, namely classifier guidance\footnote{Classifier guidance relies on a noisy classifier rather than the classifier-under-protection. See Sec.~\ref{sec:classifier-guidance} for more details.}~\cite{song2021scorebased} and classifier-free guidance~\cite{ho2021classifierfree}. If the guidance cannot match the semantics of the classifier's output, the synthesis result may fail to highlight the semantic mismatch of OODs. To address this problem, we propose several techniques that effectively utilize information from the classifier-under-protection. Additionally, we propose several test-time enhancement techniques to balance the guidance between the input image and the label condition during generation, without even fine-tuning the diffusion model. 

We evaluate the effectiveness of the proposed framework on the standard benchmark, OpenOOD~\cite{yang2022openood}. Given \cifar or \imagenet as the InD dataset, \diffguard outperforms or is on par with existing OOD detection solutions, and it can be easily combined with them to achieve state-of-the-art (SOTA) performance. We summarize the contributions of this paper as follows:
\begin{list}{\labelitemi}{\leftmargin=1em}
\setlength{\topmargin}{0pt}
\setlength{\itemsep}{0em}
\setlength{\parskip}{0pt}
\setlength{\parsep}{0pt}
    \item We propose a diffusion-based framework for detecting OODs, which directly models the semantic mismatch of OOD samples, and it is applicable to \imagenet-scale datasets;
    \item We propose several test-time techniques to improve the effectiveness of conditioning in OOD detection. Our framework can work with any pre-trained diffusion models without the need for fine-tuning, and can provide plug-and-play OOD detection capability for any classifier;
    \item Experimental results show that our framework achieves SOTA performance on \cifar and demonstrates strong differentiation ability on hard OOD samples of \imagenet. %
\end{list}

The rest of the paper is organized as follows. Section~\ref{sec:2.related} introduces related OOD detection methods and diffusion models. Section~\ref{sec:3.method} presents our framework and details the proposed solution. Experimental results are presented in Section~\ref{sec:4.exp}. We also provide discussion on limitations and future works in Section~\ref{sec:limiations}. Finally, we conclude this paper in Section~\ref{sec:5.conclusion}. 

\section{Related Work}\label{sec:2.related}

This section begins by surveying existing OOD detection methods. Especially, we demonstrate diffusion models for OOD detection, and talk about the differences between our method and other reconstruction-based ones.

\textbf{OOD Detection Methods.}
In general, OOD detection methods can be categorized as classification-based or generation-based.

Classification-based methods utilize the output from the classifier-under-protection to differentiate between OODs and InDs. 
For methods that do not modify the classifier, ODIN~\cite{liang2018enhancing}, ViM~\cite{haoqi2022vim}, MLS~\cite{pmlr-v162-hendrycks22a}, and KNN~\cite{sun2022out} are typical ones. They extract and utilize information in the feature space (e.g., KNN), the logits space (e.g., MLS, ODIN), or both (e.g., ViM). Other methods modify the classifier by proposing new losses~\cite{hsu2020generalized} or data augmentation techniques~\cite{sricharan2018building, mirzaei2023fake}, or using self-supervised training~\cite{sehwagssd, tack2020csi}.

Generation-based methods typically have a wider range of applications than classification-based ones because they have no restriction on classifiers. Most generation-based methods focus on either reconstruction quality based on inputs~\cite{denouden2018improving, schlegl2017unsupervised} or likelihood/data-density estimated from the generative model~\cite{choi2018waic,Serrà2020Input}. Their basic assumption is that generative models trained with InD data may fail to make high-quality reconstructions~\cite{denouden2018improving} or project OODs to low-density areas of the latent space~\cite{pidhorskyi2018generative}. However, this assumption may not hold true~\cite{nalisnick2018do,kirichenko2020normalizing}. In contrast, conditional synthesis does not rely on such an assumption. it differentiates OODs by constructing semantic mismatch (e.g.,~\cite{yang2022out} uses cGAN). Since semantic mismatch is the most significant property of OODs, this kind outperforms reconstruction-based ones.

Our method leverages conditional image synthesis, which shares the same benefits as \cite{yang2022out}. However, our method outperforms cGAN in terms of model training. \diffguard is compatible with any normally trained diffusion models, which eliminates the need for additional training process.

\textbf{Diffusion Models.}
Following a forward transformation from the image distribution to the Gaussian noise distribution, diffusion models~\cite{ho2020denoising} are generative models trained to learn the reverse denoising process.
The process can be either a Markov~\cite{ho2020denoising} or a non-Markov process~\cite{song2021denoising}.
The recently proposed Latent Diffusion Model (LDM)~\cite{Rombach_2022_CVPR} is a special kind. LDM conducts the diffusion process in a latent space to make the model more efficient.

\textbf{Diffusion Models for OOD Detection.}
Previously, researchers primarily utilize the reconstruction ability of diffusion models for detecting OOD and novelty instances. They achieve this by measuring the discrepancy between the input image and its reconstructed counterpart. For example, \cite{mirzaei2023fake} trains a binary classifier with training data generated from the diffusion models to differentiate OODs. \cite{liu2023outofdistribution} conducts noise augmentations to input images and then compares the differences between the denoised images and the inputs for OOD detection. Similarly, \cite{graham2022denoising} also uses diffusion models in a reconstruction-based manner, establishing a range of noise levels for the addition and removal of noise.

\begin{figure*}[tb]
    \centering
    \includegraphics[width=0.8\linewidth]{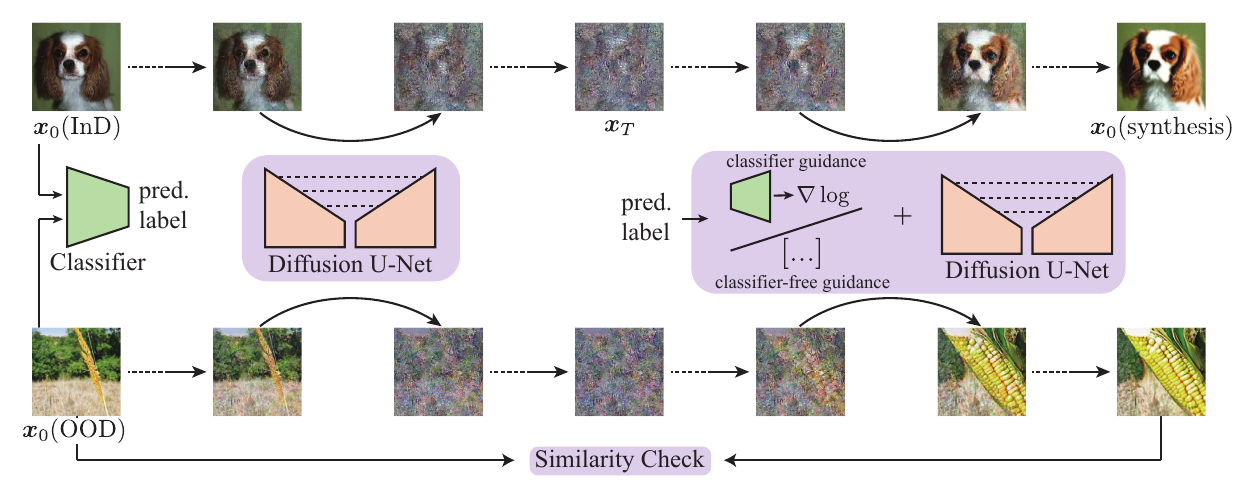}
    \caption{An overview of the \diffguard framework with diffusion models. We first use DDIM inversion to get the latent embedding ($\boldsymbol{x}_{T}$) of the input ($\boldsymbol{x}_{0}$ left). Then, we apply conditional image synthesis towards the label predicted by the classifier-under-protection. Finally, we differentiate OODs based on the similarity between the input and the synthesis. Both classifier guidance and classifier-free guidance can be applied to this framework.}
    \label{fig:overview}
\end{figure*}

Although reconstruction is one of the functions of diffusion models, a more significant advantage of diffusion models is their flexibility to handle different conditions.
Our paper employs diffusion models in detecting OODs with semantic mismatch. By utilizing both input images and semantic labels as conditions for generation, diffusion models highlight the semantic mismatch on OODs, thus facilitating the differentiation of OODs from InDs.

\section{Method}\label{sec:3.method}
In this section, we first demonstrate some preliminaries about diffusion models. Then, we present our \diffguard, which uses diffusion models for OOD detection.

\subsection{Preliminaries}
Our method is based on three significant techniques: classifier-guidance~\cite{song2021scorebased}, classifier-free guidance~\cite{ho2021classifierfree}, and DDIM inversion~\cite{song2021denoising}. The first two pertain to label conditioning methods in diffusion, whereas the last one is associated with image conditioning. We provide a concise overview of these techniques.

\textbf{Conditional Diffusion Models.}
As a member of generative models, diffusion models generate images ($\boldsymbol{x}_{0}$) through a multi-step denoising (reverse) process starting from Gaussian noise ($\boldsymbol{x}_{T}$). This process was first formulated as a Markov process by Ho \etal~\cite{ho2020denoising} with the following forward (diffusion) process:
\begin{equation}
\begin{matrix}
q(\boldsymbol{x}_{1:T}|\boldsymbol{x}_{0}):=\prod_{t=1}^{T}q(\boldsymbol{x}_{t}|\boldsymbol{x}_{t-1})
\end{matrix}
\end{equation}
where
\begin{equation}
q(\boldsymbol{x}_{t}|\boldsymbol{x}_{t-1}):=\mathcal{N}\left(\sqrt{\frac{\alpha_{t}}{\alpha_{t-1}}}\boldsymbol{x}_{t-1},(1-\frac{\alpha_{t}}{\alpha_{t-1}})\boldsymbol{I}\right)
\end{equation}
and the decreasing sequence $\alpha_{1:T}\in(0,1]^{T}$ is the transition coefficient.
After refactoring the process to be non-Markov, Song \etal~\cite{song2021denoising} proposed a skip-step sampling strategy to speedup the generation, as in Eq.~(\ref{eq:ddim-general}),
where $t\in[1,...,T]$, $\epsilon_{t}\sim\mathcal{N}(\mathbf{0},I)$ is the standard Gaussian noise independent of $\boldsymbol{x}_{t}$, and $\epsilon_{\theta}^{(t)}$ is the estimated noise by the model $\theta$ at timestep $t$. The sampling process can be performed on any sub-sequence $t\in\tau\subset[1,...,T]$.
\begin{equation}
    \begin{split}
        \boldsymbol{x}_{t-1} &=
        \sqrt{\alpha_{t-1}}\Big(\frac{\boldsymbol{x}_{t}-\sqrt{1-\alpha_{t}}\epsilon_{\theta}^{(t)}(\boldsymbol{x}_{t})}{\sqrt{\alpha_{t}}}\Big)\\
        &+\sqrt{1-\alpha_{t-1}-\sigma_{t}^{2}}\cdot\epsilon_{\theta}^{(t)}(\boldsymbol{x}_{t}) + \sigma_{t}\epsilon_{t}%
    \end{split}
    \label{eq:ddim-general}
\end{equation}

Under this formulation, there are two ways to apply the label semantic condition $\boldsymbol{y}$ to the generation process: classifier guidance and classifier-free guidance. For classifier guidance~\cite{song2021scorebased,dhariwal2021diffusion}, the condition-guided noise prediction $\hat{\epsilon}(\boldsymbol{x}_{t})$ is given by (we omit $\theta$ and $t$ in $\epsilon(\cdot)$):
\begin{equation}
    \hat{\epsilon}(\boldsymbol{x}_{t}):=\epsilon(\boldsymbol{x}_{t}) + s\sqrt{1-\alpha_{t}}\cdot\nabla_{\boldsymbol{x}_{t}}\log p_{\phi}(\boldsymbol{y}|\boldsymbol{x}_{t})\text{,}
    \label{eq:classifier-guidance}
\end{equation}
where $\log p_{\phi}$ is given by a classifier trained on noisy data $\boldsymbol{x}_{t}$, and $s$ is to adjust the guidance scale (\ie, strength of the guidance).
For classifier-free guidance~\cite{ho2021classifierfree,nichol22glide}, a conditional diffusion model $\bar{\epsilon}_{\theta}^{(t)}(\boldsymbol{x}_{t}, \boldsymbol{y})$ is trained. The training objective is the same as vanilla diffusion models, but $\epsilon$ changes to $\tilde{\epsilon}$ during inference as follows (we omit $\theta$ and $t$):
\begin{equation}
    \tilde{\epsilon}(\boldsymbol{x}_{t},\boldsymbol{y}):=\bar{\epsilon}(\boldsymbol{x}_{t}, \emptyset) + \omega[{\bar{\epsilon}(\boldsymbol{x}_{t}, \boldsymbol{y}) - \bar{\epsilon}(\boldsymbol{x}_{t}, \emptyset)}]\text{,}
    \label{eq:classifier-free-guidance}
\end{equation}
where $\omega$ is to adjust the guidance scale. Both classifier guidance and classifier-free guidance are qualified for conditional generation.

\textbf{The Inversion Problem of Diffusion Models.}
For generative models, applying the input image as a condition for synthesis can be done by solving the inversion problem~\cite{xia2022gan}.
By applying score matching~\cite{NEURIPS2019_3001ef25} to the formulated SDE, the diffusion process can be converted into an Ordinary Differential Equation (ODE)~\cite{song2021scorebased}, which provides a deterministic transformation between an image and its latent. This is also applied to the inference process of DDIM (where $\sigma=0$ in Eq.~\eqref{eq:ddim-general}). Thus, the diffusion process from an image ($\boldsymbol{x}_{0}$) to its latent ($\boldsymbol{x}_{T}$) is given by: %
\begin{equation}
    \begin{split}
        \boldsymbol{x}_{t+1} &=
        \sqrt{\alpha_{t+1}}\Big(\frac{\boldsymbol{x}_{t}-\sqrt{1-\alpha_{t}}\epsilon(\boldsymbol{x}_{t})}{\sqrt{\alpha_{t}}}\Big)\\
        &+\sqrt{1-\alpha_{t+1}}\epsilon(\boldsymbol{x}_{t})\text{, where }t\in[0,...,T-1].
    \end{split}\label{eq:6}
\end{equation}
Such a latent can be used to reconstruct the input through the denoising process.
\subsection{Diffusion Models for OOD Detection}
We show an overview of the proposed framework in Fig.~\ref{fig:overview}. Given a classifier-under-protection, we utilize its prediction of the input and synthesize a new image conditioned on both the predicted label and the input. Intuitively, if the predicted label does not match the input (i.e., OOD), dissimilarity will be evident between the synthesis and the input, and vice versa. Then, we can assess whether an input image is OOD by evaluating the similarity between the input and its corresponding synthesis.

For the two conditions, the label condition tends to change the content to reflect its semantics while the input image condition tends to keep the synthesis as original.
Therefore, the main challenge of our method is to apply and balance the two conditions.
To handle the input image as a condition, diffusion models' inversion ability (e.g., DDIM~\cite{song2021denoising}) serves as an advantage in faithfully restoring the contents.
For the label condition, since there are two fundamentally different methods in diffusion, namely classifier guidance and classifier-free guidance, we propose different techniques for them to better differentiate OODs.
We demonstrate the proposed methods respectively in the following of this section.

\subsubsection{Diffusion with Classifier Guidance}\label{sec:classifier-guidance}
\begin{figure}[t]
    \centering
    \includegraphics[width=0.9\linewidth]{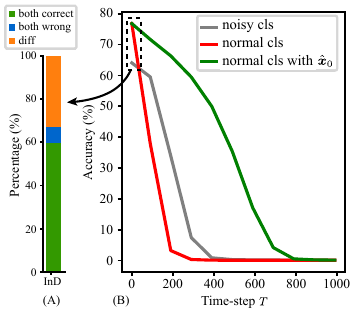}
    \caption{Different behavior between a noisy classifier and a normal ResNet50 classifier on \imagenet validation. (A) Conflicting predictions: nearly 35\% of the predictions are different; (B) The accuracy degradation throughout the diffusion process.}
    \label{fig:acc_with_grad}
\end{figure}

In classifier-guided diffusion models, the classifier trained on noisy data is the key to conditional generation, as shown in Eq.~\eqref{eq:classifier-guidance}.
However, directly using a classifier trained on such data for OOD detection is problematic.
With a different training process, the classifier may predict differently from the classifier-under-protection, even on clean samples. As shown in Fig.~\ref{fig:acc_with_grad} (A), when using a ResNet50 as the classifier-under-protection, differences in prediction exist in nearly 35\% of the image samples.

The problem above hinders us from using a noisy classifier for guidance. In this section, we replace the noisy classifier $\phi$ with the exact classifier-under-protection $\phi_{n}$. Then, we propose two techniques for better utilization of the classifier for OOD detection.

\begin{figure}[t]
    \centering
    \includegraphics[width=0.9\linewidth]{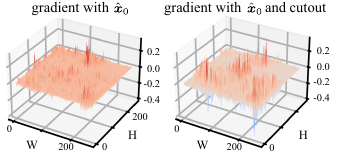}
    \caption{Gradient visualizations of classifier-guided diffusion with (right) and without (left) cutout at $t=600$. We use a normal ResNet50 classifier from \imagenet.}
    \label{fig:acc_with_grad_c}
\end{figure}

\textbf{Tech \#1: Clean Grad: using the gradient from a normal classifier.}
At the right-hand side of Eq.~\eqref{eq:ddim-general}, the first term can be interpreted as an estimation of $\boldsymbol{x}_0$, i.e., $\hat{\boldsymbol{x}}_{0} = \frac{\boldsymbol{x}_{t}-\sqrt{1-\alpha_{t}}\epsilon_{\theta}^{(t)}(\boldsymbol{x}_{t})}{\sqrt{\alpha_{t}}}$.
In this case, we can use $\hat{\boldsymbol{x}}_{0}$ as a substitute of $\boldsymbol{x}_{t}$. Calculation of the gradient on $\boldsymbol{x}_{t}$ in Eq.~\eqref{eq:classifier-guidance} can be transformed into that on $\hat{\boldsymbol{x}}_{0}$, shown as follows:
\begin{equation}
\nabla_{\boldsymbol{x}_{t}}\log p_{\phi}(y|\boldsymbol{x}_{t}):=\nabla_{\boldsymbol{x}_{t}}\log p_{\phi_{n}}(y|\hat{\boldsymbol{x}}_{0}(\boldsymbol{x}_{t}))\text{.}
\end{equation}
With such an $\hat{\boldsymbol{x}}_{0}$ as input, the classifier can provide a correct gradient of log-probability for a wide range of $t$, thus offering more accurate generation directions and leading to better semantic guidance.

To understand the operation, we plot the changes in classification accuracy with different time-steps $t$, shown in Fig.~\ref{fig:acc_with_grad} (B).
The classification accuracy reflects the prediction quality of $\log p$, and thus the quality of $\nabla\log p$.
With the noisy $\boldsymbol{x}_{t}$ as input, the accuracy of the normal classifier degrades more dramatically than the noisy classifier. However, with $\hat{\boldsymbol{x}}_{0}$ as input, the classification accuracy of the normal classifier reduces much slower than the other two cases.

Besides, we propose that data augmentation is important to successfully applying a normal classifier.
Using $\hat{\boldsymbol{x}}_{0}$, the gradient of a normal classifier is relatively small and flat, as shown in Fig.~\ref{fig:acc_with_grad_c} left.
Since gradient is the only term representing the direction of semantics in Eq.~\eqref{eq:classifier-guidance}, it is hard for a flat gradient to effectively change the semantics of the image during synthesis.
To solve this problem, we propose to use data augmentations (\ie, random cutout) as follows:
\begin{equation}
\small
\nabla_{\boldsymbol{x}_{t}}\log p_{\phi}(y|\boldsymbol{x}_{t}):=\nabla_{\boldsymbol{x}_{t}}%
\log p_{\phi_{n}}(y|\operatorname{cutout}(\hat{\boldsymbol{x}}_{0}(\boldsymbol{x}_{t}))).\label{eq:clean grad}
\end{equation}
On the one hand, the gradient with augmentations is sharper and with higher amplitude (shown in Fig.~\ref{fig:acc_with_grad_c} right), which is better for effective semantic changes on the image than that without augmentations.
On the other hand, gradients corresponding to different augmentations can be accumulated to form more comprehensive guidance.
To better interpret the effect, we provide a qualitative ablation study in Sec.~\ref{sec:ablation}.

\textbf{Tech \#2: Adaptive Early-Stop (AES) of the diffusion process.}
From Fig.~\ref{fig:acc_with_grad} (B), we notice that both classifiers experience a sharp accuracy drop with increasing $t$. 
This reminds us that there exists a $t_{stop}$ such that the classifier cannot provide meaningful semantic guidance when $t>t_{stop}$.
Therefore, it is necessary to apply early-stop when performing image inversion.

Instead of setting a fixed step to stop, we propose to adaptively stop the inversion process according to the quality of the diffused image. Specifically, we use distance metrics (e.g., Peak Signal-to-Noise Ratio (PSNR) and DISTS~\cite{ding2020iqa}) to measure the pixel-level difference between the diffused image and the corresponding image input. If the quality degradation exceeds a given threshold, we stop the diffusion and start the synthesis (denoising) process. Empirically, such a threshold is located around $t=600=3/5T$, as evidenced from Fig.~\ref{fig:acc_with_grad} (B).

The principle of using the adaptive manner of early-stop lies in the trade-off between consistency and controllability.
The early-stop technique is adopted in several literatures~\cite{kwon2023diffusion,liew2022magicmix}, as image inversion through DDIM occasionally fails to guarantee a coherent reconstruction.
Specifically, fewer inversion/generation steps lead to better consistency but lower controllability, and vice versa~\cite{meng2022sdedit}.
For example, LPIPS is used in \cite{kwon2023diffusion} as a measure to balance image editing strength and generation quality. 

\begin{figure}[tb]
    \centering
    \includegraphics[width=\linewidth]{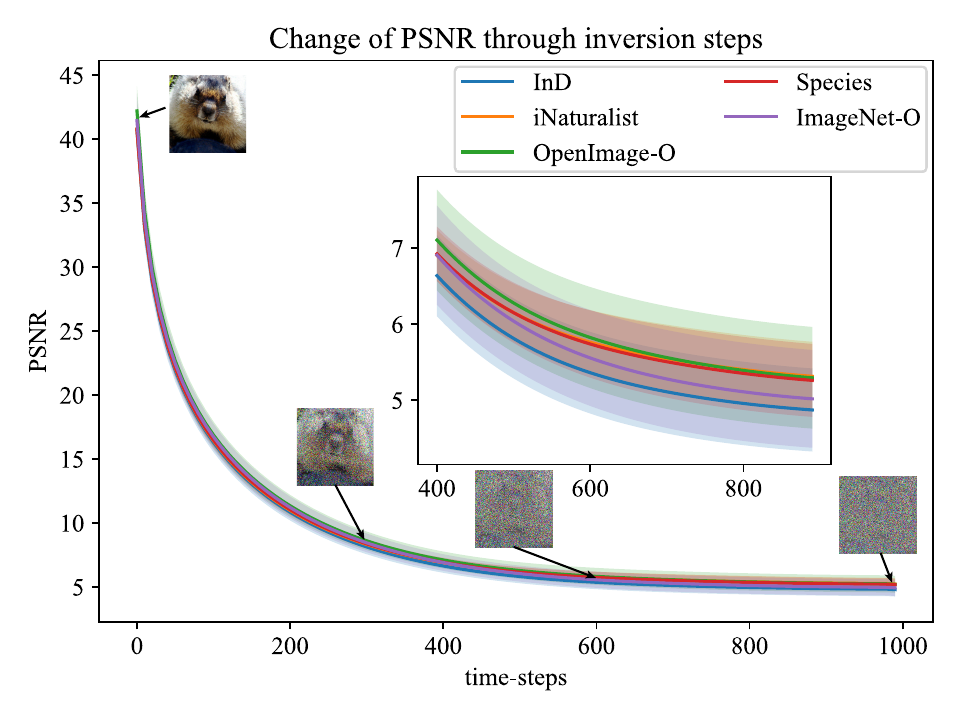}
    \caption{PSNR changes throughout the diffusion process. The data is collected from the \imagenet validation set and 4 OOD datasets, with a ResNet50 classifier.}
    \label{fig:m_changes}
\end{figure}

For OOD detection tasks, we observe that InD and OOD samples have different patterns of quality degradation through the inversion process, especially reflected by PSNR and DISTS. Fig.~\ref{fig:m_changes} shows such a phenomenon. The empirical fact that InD data has faster quality degradation rates than OOD data acts as a good property to monitor the diffusion process. As a result, we can set a proper threshold with different purposes for InD and OOD samples. The threshold generally corresponds to fewer diffusion steps on InD samples, ensuring faithful reconstruction. Simultaneously, it also leads to greater steps on OOD samples, ensuring better controllability towards label conditions, and thus more significant differences compared with the inputs.

\subsubsection{Diffusion with Classifier-Free Guidance}\label{sec:classifier-free-guidance}
Classifier-free guidance~\cite{ho2021classifierfree} relies on a trained conditional diffusion model.
Benefiting from the conditional training process, it is not necessary to further apply an external classifier. In addition, the attention-based condition injection results in better coherence between the synthesis and the given label condition~\cite{nichol22glide}.
However, we find that the guidance scale ($\omega$ in Eq.~\eqref{eq:classifier-free-guidance}) of the condition is a double-edged sword. For semantic mismatch, we rely on the differences between the syntheses given consistent and inconsistent conditions. A small guidance scale cannot provide semantic changes large enough to drive the OOD samples towards the inconsistent predictions, while a large guidance scale drastically changes both InD and OOD samples, also increasing the difficulty in differentiation. Therefore, it is critical to reach a trade-off with this single parameter.

\begin{figure}[t]
    \centering
    \includegraphics[width=0.8\linewidth]{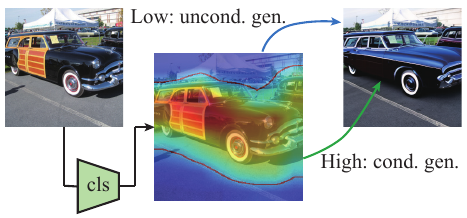}
    \caption{Illustration of the classifier-free guidance with CAM. CAM helps to utilize information given by the classifier-under-protection. For areas with high activation, we conduct label-guided generation; for areas with low activation, we drop the label guidance and perform the original DDIM-based reconstruction.}
    \label{fig:cam_gen}
\end{figure}

\textbf{Tech \#3: Distinct Semantic Guidance (DSG).} To solve the issue stated above, we apply the Class Activation Map (CAM~\cite{selvaraju2017grad}) of the classifier-under-protection to impose restrictions on the generation process.
Specifically, we apply classifier-free guidance to high-activation areas while applying the vanilla unconditional generation to other areas, with the procedure shown in Fig.~\ref{fig:cam_gen}.

While using masks to guide the generation process differently has been a common practice in image editing~\cite{nichol22glide,hertz2022prompt}, CAM in DSG associates the label guidance with the classifier-under-protection, which provides crucial information to construct and highlight semantic mismatch.

According to the CAM, high-activation areas are crucial to prediction, thus having a direct correlation with the predicted label, and vice versa.
For InD samples, applying the guidance to high-activation areas effectively limits its scope of effect, thus mitigating unwanted distortions; for OOD cases, guidance on these areas leads to high inconsistency, as they are forced to embed target semantics that they do not originally have.
As a result, we can easily differentiate OOD cases by similarity measurements.

\section{Experiments}\label{sec:4.exp}
\subsection{Experimental Setups}
\textbf{Benchmarks and Datasets.} 
We evaluate \diffguard following a widely adopted semantic OOD detection benchmark, OpenOOD~\cite{yang2022openood}.
OpenOOD unifies technical details in evaluation (e.g., image pre-processing procedures and classifiers) and proposes a set of evaluation protocols for OOD detection methods.
For each InD dataset, it categorizes OODs into different types (i.e., near-OOD and far-OOD) for detailed analyses.

In this paper, we employ \cifar~\cite{krizhevsky2009learning} and \imagenet~\cite{krizhevsky2017imagenet} as InD samples, respectively.
\cifar is mostly adopted for evaluation, though being small in scale. 
We choose near-OODs (i.e. \cifarh~\cite{krizhevsky2009learning} and \tinyimagenet~\cite{le2015tiny}).
For large-scale evaluation, we set \imagenet as InD. OODs are also selected from the near-OODs in OpenOOD: Species~\cite{pmlr-v162-hendrycks22a}, iNaturalist~\cite{huang2021mos}, ImageNet-O~\cite{hendrycks2021natural}, and OpenImage-O~\cite{haoqi2022vim}.

\textbf{Metrics.} Following OpenOOD, we adopt the Area Under Receiver Operating Characteristic curve (AUROC) as the main metric for evaluation. AUROC reflects the overall detection capability of a detector.
Besides, we consider FPR@95, which evaluates the False Positive Rate (FPR) at 95\% True Positive Rate (TPR). The widely used 95\% TPR effectively reflects the performance in practice.

\textbf{Baselines.} For comparison, we consider two types of baselines.
For classification-based methods, we involve recently proposed well-performing methods, including EBO~\cite{liu2020energy}, KNN~\cite{sun2022out}, MLS~\cite{pmlr-v162-hendrycks22a} and ViM~\cite{haoqi2022vim}.
EBO uses energy-based functions based on the output from the classifier.
KNN performs OOD detection by calculating the non-parametric nearest-neighbor distance.
MLS identifies the value of the maximum logit scores (without softmax). ViM proposes to combine logits with features for OOD detection. All of them are strong baselines according to OpenOOD.
For generation-based methods, we consider a recent method, DiffNB~\cite{liu2023outofdistribution}, which utilizes the denoising ability of diffusion and performs reconstruction within the neighborhood.
Following OpenOOD, all the classification-based baselines work on both \cifar and \imagenet. For DiffNB, we use the official implementation
and only compare with it on \cifar.

\begin{table}[t]
    \footnotesize{\begin{center} 
    \setlength\tabcolsep{0.8pt}
    \begin{tabular}{c|cc|cc|cc}
      \toprule
      \multirow{3}{*}{Method} & \multicolumn{2}{c|}{\cifarh} & \multicolumn{2}{c|}{\tinyimagenet} & \multicolumn{2}{c}{average} \\ \cline{2-7} 
                & AUROC & FPR@95
                & AUROC & FPR@95
                & AUROC & FPR@95 \\
                & $\uparrow$ & $\downarrow$
                & $\uparrow$ & $\downarrow$
                & $\uparrow$ & $\downarrow$ \\
      \midrule
      EBO~\cite{liu2020energy}
                & 86.19 & \uit{51.32}
                & 88.61 & \uit{44.89}
                & 87.41 & \uit{48.11}  \\
      KNN~\cite{sun2022out}
            & 89.62 & 52.19
            & 91.48 & 46.18
            & 90.55 & 49.19 \\
      MLS\cite{pmlr-v162-hendrycks22a} 
                & 86.14 & 52.04
                & 88.53 & 45.38
                & 87.34 & 48.71 \\
      ViM\cite{haoqi2022vim}
                & 87.16 & 56.81 
                & 88.85 & 52.89 
                & 88.01 & 54.85  \\
      MC-Dropout\cite{pmlr-v48-gal16}
                & 86.74 & 61.49
                & 88.32 & 58.44
                & 87.53 & 59.97  \\
      Deep Ens.\cite{pmlr-v70-guo17a}
            & 89.97 & 54.61
            & 91.31 & 51.23
            & 90.64 & 52.92 \\
      ConfidNet$^{*}$\cite{corbiere2019addressing}
                & 85.92 & 72.37
                & 87.16 & 69.75
                & 86.54 & 71.06 \\
      \hline
      DiffNB~\cite{liu2023outofdistribution} 
                & 89.79 & 53.23
                & 91.77 & 45.88
                & 90.78 & 49.56 \\
      \midrule
      Ours      & 89.88 & 52.67
                & 91.88 & 45.48
                & 90.88 & 49.08 \\
      Ours\tiny{+EBO}      
                & \uit{89.93} & \textbf{50.77}
                & \uit{91.95} & \textbf{43.58}
                & \uit{90.94} & \textbf{47.18} \\
      Ours\tiny{+Deep Ens.}      
                & \textbf{90.40} & 52.51
                & \textbf{91.98} & 45.04
                & \textbf{91.19} & 48.78 \\
      Ours(Oracle)
                & 98.34 & 7.94
                & 98.52 & 7.11
                & 98.43 & 7.53 \\
      \bottomrule
    \end{tabular}
    \end{center}}
    \caption{The OOD detection performance with \cifar as InD. The diffusion model uses classifier-free guidance. All the values are in percentages. $\uparrow/\downarrow$ indicates that a higher/lower value is better. The best results are in \textbf{bold}, and the second best results are in \uit{underlined italic}. (Oracle) indicates we use an Oracle InD classifier. $^{*}$ use VGG16 classifier.}
    \label{tab:cifar}
\end{table}

\textbf{Diffusion Models.} To evaluate the OOD detection ability of \diffguard, we consider different types of diffusion models. Specifically, on \cifar, we use the same pre-trained model as DiffNB, which is a conditional DDPM~\cite{ho2020denoising} with classifier-free guidance. On \imagenet, we use the unconditional Guided Diffusion Model (GDM)~\cite{dhariwal2021diffusion} and apply classifier guidance.
GDM is an advanced version of DDPM~\cite{ho2020denoising} with optimizations on the architecture and the model training process. Besides, we also adopt the Latent Diffusion Model (LDM)~\cite{Rombach_2022_CVPR}
as an example of classifier-free guidance. As stated in Sec.~\ref{sec:2.related}. LDM is a prevailing diffusion model in text-guided image generation~\cite{zhang2023adding} due to its efficient architecture.

\textbf{Classifiers-under-protection.} We directly apply the off-the-shelf settings of classifiers-under-protection established by OpenOOD.
Specifically, we use ResNet18~\cite{he2016deep} trained on \cifar and ResNet50~\cite{he2016deep} trained on \imagenet. The pre-trained weights for both can be found in OpenOOD's GitHub\footnote{https://github.com/Jingkang50/OpenOOD}.

\begin{table*}[t]
    \small{\begin{center} 
        \setlength\tabcolsep{1.5pt}
        \begin{tabular}{c|cc|cc|cc|cc|cc}
        \toprule
        \multirow{2}{*}{Method} &
          \multicolumn{2}{c|}{Species} &
          \multicolumn{2}{c|}{iNaturalist} &
          \multicolumn{2}{c|}{OpenImage-O} &
          \multicolumn{2}{c|}{ImageNet-O} &
          \multicolumn{2}{c}{Average} \\
        \cline{2-11} 
                           & AUROC \textcolor{red}{$\uparrow$} & FPR@95 \textcolor{green}{$\downarrow$} 
                           & AUROC \textcolor{red}{$\uparrow$} & FPR@95 \textcolor{green}{$\downarrow$}
                           & AUROC \textcolor{red}{$\uparrow$} & FPR@95 \textcolor{green}{$\downarrow$}
                           & AUROC \textcolor{red}{$\uparrow$} & FPR@95 \textcolor{green}{$\downarrow$}
                           & AUROC \textcolor{red}{$\uparrow$} & FPR@95 \textcolor{green}{$\downarrow$} \\
        \midrule
        EBO~\cite{liu2020energy} & 72.04 & 82.33 & 90.61 & 53.83 & 89.15 & 57.10 & 41.91 & 100.00 & 73.43 & 73.31  \\
        KNN~\cite{sun2022out}
            & 76.38 & 76.19
            & 85.12 & 68.41
            & 86.45 & 57.56
            & 75.37 & 84.65
            & 80.83 & 71.70 \\
        ViM~\cite{haoqi2022vim}
            & 70.68 & 83.94
            & 88.40 & 67.85
            & 89.63 & 57.56
            & 70.88 & 85.30
            & 79.90 & 73.66  \\
        MLS~\cite{pmlr-v162-hendrycks22a}
            & 72.89 & 80.87
            & 91.15 & 50.80
            & 89.26 & 57.11
            & 40.85 & 100.00
            & 73.54 & 72.20   \\
        \hline
        Ours(GDM)   & 73.19\tiny{$\pm$0.18} & 83.68\tiny{$\pm$0.22}
                    & 85.81\tiny{$\pm$0.16} & 71.23\tiny{$\pm$0.54}
                    & 82.32\tiny{$\pm$0.30} & 74.80\tiny{$\pm$0.38}
                    & 65.23\tiny{$\pm$0.19} & 87.74\tiny{$\pm$0.20}
                    & 76.64\tiny{$\pm$0.13} & 79.36\tiny{$\pm$0.12}  \\
        Ours(LDM)     & 65.87 & 91.70
                      & 75.64 & 79.06
                      & 73.92 & 81.19
                      & 68.57 & 84.35
                      & 71.00 & 84.08 \\
        Ours(GDM)\tiny{+KNN}
            &	\textbf{77.81}\tiny{\textcolor{red}{+1.43}}& 71.04\tiny{\textcolor{green}{-5.15}}
            &	90.19\tiny{\textcolor{red}{+5.07}}&	48.79\tiny{\textcolor{green}{-19.62}}
            &	87.80\tiny{\textcolor{red}{+1.35}}&	52.80\tiny{\textcolor{green}{-4.76}}
            &	\textbf{75.68}\tiny{\textcolor{red}{+0.31}}& \textbf{80.85}\tiny{\textcolor{green}{-3.80}}
            &	\textbf{82.87}\tiny{\textcolor{red}{+2.04}}& 63.37\tiny{\textcolor{green}{-8.33}} \\
        Ours(GDM)\tiny{+ViM}
            &	74.48\tiny{\textcolor{red}{+3.80}}&	72.26\tiny{\textcolor{green}{-11.68}}
            &	92.50\tiny{\textcolor{red}{+4.10}}&	39.09\tiny{\textcolor{green}{-28.76}}
            &	\textbf{91.11}\tiny{\textcolor{red}{+1.48}}& 45.02\tiny{\textcolor{green}{-12.54}}
            &	72.42\tiny{\textcolor{red}{+1.54}}&	82.30\tiny{\textcolor{green}{-3.00}}
            &	82.63\tiny{\textcolor{red}{+2.73}}&	59.67\tiny{\textcolor{green}{-14.00}} \\
        Ours(LDM)\tiny{+ViM} & 71.08\tiny{\textcolor{red}{+0.40}} & 82.20\tiny{\textcolor{green}{-1.74}}       
                             & 89.39\tiny{\textcolor{red}{+0.99}} & 61.01\tiny{\textcolor{green}{-6.84}}       
                             & 89.65\tiny{\textcolor{red}{+0.02}} & 55.83\tiny{\textcolor{green}{-1.73}}       
                             & 74.85\tiny{\textcolor{red}{+3.97}} & 81.95\tiny{\textcolor{green}{-3.35}}       
                             & 81.24\tiny{\textcolor{red}{+1.35}} & 70.25\tiny{\textcolor{green}{-3.41}}       \\
        Ours(GDM)\tiny{+MLS}
            &	75.95\tiny{\textcolor{red}{+3.06}}&	\textbf{70.31}\tiny{\textcolor{green}{-10.56}}
            &	\textbf{93.03}\tiny{\textcolor{red}{+1.88}}& \textbf{30.74}\tiny{\textcolor{green}{-20.06}}
            &	90.74\tiny{\textcolor{red}{+1.48}}&	\textbf{40.61}\tiny{\textcolor{green}{-16.50}}
            &	65.72\tiny{\textcolor{red}{+24.87}}& 87.05\tiny{\textcolor{green}{-12.95}}
            &	81.36\tiny{\textcolor{red}{+7.82}}&	\textbf{57.18}\tiny{\textcolor{green}{-15.02}} \\
        Ours(LDM)\tiny{+MLS} & 73.69\tiny{\textcolor{red}{+0.80}}   & 75.91\tiny{\textcolor{green}{-4.96}}       
                             & 91.55\tiny{\textcolor{red}{+0.40}}   & 43.56\tiny{\textcolor{green}{-7.24}}       
                             & 89.61\tiny{\textcolor{red}{+0.35}}  & 50.61\tiny{\textcolor{green}{-6.50}}       
                             & 69.33\tiny{\textcolor{red}{+28.48}} & 84.00\tiny{\textcolor{green}{-16.00}}       
                             & 81.05\tiny{\textcolor{red}{+7.51}}  & 63.52\tiny{\textcolor{green}{-8.68}}   \\
        \bottomrule
    \end{tabular}
    \end{center}}
    \caption{The OOD detection performance with \imagenet as InD. GDM uses classifier guidance, while LDM uses classifier-free guidance. All the values are in percentages. $\textcolor{red}{\uparrow}/\textcolor{green}{\downarrow}$ indicates that a higher/lower value is better. The best results are in \textbf{bold}. We highlight the comparisons with colors when combining \diffguard with other baselines. For AUROC with Ours(GDM), we present the average and standard deviation over four runs. There is no randomness in LDM.}
    \label{tab:imagenet}
\end{table*}

\textbf{Similarity Metric.} For simplicity, we use generic similarity metrics across different diffusion models and different OODs. Specifically, we choose $\ell_{1}$ distance on logits between input image and its synthetic counterpart for \cifar benchmark, as in \cite{liu2023outofdistribution}; choose DISTS distance~\cite{kastryulin2022piq} for \imagenet benchmark (except Table~\ref{tab:early-stop}).

Note that all diffusion models are pre-trained only with InD data. We do not fine-tune them. For more implementation details, please refer to the supplementary material.

\subsection{Results on \cifar}\label{sec:res-on-cifar}

Table~\ref{tab:cifar} shows the results on \cifar. \diffguard outperforms or at least is on par with other methods on these two near-OOD datasets.
In terms of AUROC, \diffguard performs better than all other baselines. \diffguard inherits the merit of image space differentiation in generation-based methods, which makes it better than classification-based ones. By highlighting the semantic mismatch, it further outperforms the generation-based DiffNB even with the same diffusion model.
In terms of FPR@95, \diffguard also outperforms DiffNB.
Although classification-based methods perform slightly better than ours, we show that \diffguard can work with them to establish new SOTAs. Specifically, the combined method only trusts samples with high detection confidence by the baselines, while resorting to \diffguard for hard cases. As an example, \diffguard + Deep Ensemble performs best on AUROC and \diffguard+ EBO performs best on FPR@95 in the near-OOD benchmark for \cifar.  

Note that the semantic mismatch utilized by \diffguard comes from the predicted label and the input image. Wrong prediction from the classifier on InDs may affect the performance of the framework. To avoid such negative effects, we establish a hypothetical oracle classifier, as shown in the last row of Table~\ref{tab:cifar}. Specifically, this oracle classifier outputs the ground-truth labels for InDs and random labels for OODs. We notice both results get improved by a large margin. Especially, \diffguard can reach a 95\% TPR with very low FPRs.
In practice, such a phenomenon reminds us of a common property in OOD detection~\cite{yang2022out,haoqi2022vim}: the performance (of \diffguard) can improve with the increasing accuracy of the classifier.

\subsection{Results on \imagenet}
Table~\ref{tab:imagenet} shows the results on the \imagenet benchmark.
\imagenet is hard for OOD detection due to both its large scale and difficulty in semantic differentiation.
We investigate the ability of \diffguard in differentiating hard OOD cases. For example, on Species, none of the baselines perform well, while using GDM with \diffguard outperforms all baselines in terms of both AUROC and FPR@95.
On ImageNet-O, many baseline methods tend to assign higher scores to OODs rather than InDs, as indicated by AUROC $< 50\%$, which shows they fail to detect OODs.
However, \diffguard can still keep its performance and achieve the best FPR@95 with LDM.

We further validate the performance of \diffguard on hard samples by combining it with some classification-based methods. We use the same method as that on \cifar (stated in Sec.~\ref{sec:res-on-cifar}). The performance improvements are shown in the last 5 rows of Table~\ref{tab:imagenet}. Especially, \diffguard saves the worst-case performance of baselines. For example on ImageNet-O, \diffguard brings considerable improvement to MLS. Besides, for average performance, \diffguard helps to reach SOTA on this benchmark.

Another comparison shown in Table~\ref{tab:imagenet} is between GDM and LDM in \diffguard.
We notice that GDM performs better in general when used both alone and with other baselines, while LDM stays at a close level and sometimes has a lower FPR@95.
As long as the diffusion model can synthesize high-quality images, \diffguard can use it to detect OODs.
Beyond OOD detection performance, the choice of diffusion models can be made according to other properties.
For example, GDM has a simpler architecture~\cite{dhariwal2021diffusion}, while LDM uses fewer DDIM timesteps (as evidenced in Sec.~\ref{sec:ablation}), and thus is faster in inference.
For different techniques proposed for both classifier guidance and classifier-free guidance, we provide ablation studies to analyze their effectiveness in Sec.~\ref{sec:ablation}.

\subsection{Ablation Study}\label{sec:ablation}
In this section, we provide some in-depth analyses regarding the effectiveness of each technique proposed for \diffguard. For more qualitative analyses such as failure cases, please refer to the supplementary material.

\begin{table}[h]
    \small{
    \begin{center}    
        \setlength\tabcolsep{1pt}
        \begin{tabular}{c|cccc|c}
        \toprule
        \footnotesize{Method} &
          \footnotesize{Species} &
          \footnotesize{iNaturalist} &
          \footnotesize{OpenImage-O} &
          \footnotesize{ImageNet-O} &
          \footnotesize{Average} \\
        \hline
        \footnotesize{w/o $\hat{x}_{0}$, w/o aug}
            & 66.45 & 64.80 & 48.80 & 42.30 & 55.59 \\
        \footnotesize{only w/ aug}
            & 71.16 & 85.77 & 74.17 & 56.06 & 71.79 \\
        \footnotesize{only w/ $\hat{x}_{0}$}
            & 71.95 & 84.11 & 80.72 & 63.82 & 75.15 \\
        \hline
        Ours & 
            \textbf{73.19} &
            \textbf{85.81} &
            \textbf{82.32} &
            \textbf{65.23} &
            \textbf{76.64} \\
        \bottomrule
    \end{tabular}
    \end{center}
    }
    \caption{Ablation for Clean Grad on GDM with ImageNet as InD. We show AUROC with different OODs. The related settings are the same as in Table~2.}
    \label{tab:imagenet-abl}
\end{table}

\begin{figure}[tb]
    \centering
    \includegraphics[width=0.95\linewidth]{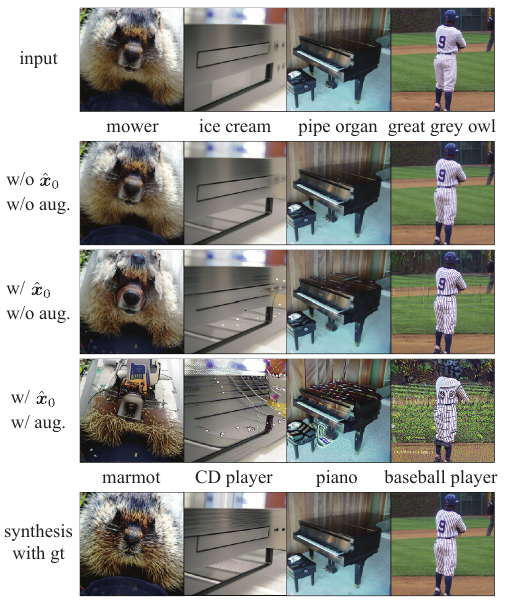}
    \caption{Ablation study to show the effectiveness of using $\hat{x}_{0}$ and data augmentations. The images are from the \imagenet validation set. There is a clear difference between the ground-truth-guided syntheses and the false-label-guided ones.}
    \label{fig:guided-ablation}
\end{figure}

\textbf{Comparisons of Clean Grad.}
We ablate the usage of either $\hat{x}_{0}$ or data augmentation in the proposed Clean Grad on classifier guidance, and show how AUROC changes in Table~\ref{tab:imagenet-abl}. As can be seen, both techniques bring significant improvements. The best results are achieved by combining them together.
Besides, we qualitatively validate their effectiveness, as shown in Fig.~\ref{fig:guided-ablation}.
For simplicity, we only show InD samples and use false-label guidance to show the effects on OODs. First, we identify the difficulty in manipulating noisy semantics with a normal classifier. Without $\hat{\boldsymbol{x}}_{0}$ or cutout, the diffusion model fails to make visually perceptible modifications.
Then, by adding either $\hat{\boldsymbol{x}}_{0}$ or cutout, we can identify differences to some extent. Finally, after applying both $\hat{x}_{0}$ and data augmentations, the generation results manage to reflect the given label.
As a comparison, the model can guarantee faithful reconstruction and negligible distortion when synthesizing with ground-truth labels (last row in Fig.~\ref{fig:guided-ablation}).
Such results show the effectiveness of our techniques in benefiting similarity measurements, and thus OOD detection.

\textbf{Early-stop Metrics in AES.}
As shown in Fig.~\ref{fig:m_changes}, early-stop contributes to the differentiation ability of \diffguard.
In Table~\ref{tab:early-stop}, we show the effect of different early-stop metrics and how AUROC varies with their thresholds.
We note that different metrics perform differently on different OODs. Specifically, PSNR performs better on Species, while DISTS performs better on the others.
In practice, we can combine them together to reach better average-case performance (as shown in the last row of Table~\ref{tab:early-stop}).

In practice, it could be easy to choose a proper threshold for each distance.
As stated in Sec.~\ref{sec:classifier-guidance}, the intuition of early-stop is to ensure meaningful semantic guidance by the classifier. Therefore, one can choose an initial threshold according to the change of the classification accuracy as shown in Fig.~\ref{fig:acc_with_grad}. Here, we pick the values at $t=3/5T$ and vary slightly to show the effectiveness, shown in Table~\ref{tab:early-stop}.

\begin{table}[t]
    \small{\begin{center}
    \setlength\tabcolsep{2pt}
    \begin{tabular}{cc|cccc|c}
    \toprule
    \footnotesize{PSNR} &
    \footnotesize{DISTS} &
                   \footnotesize{Species} & 
                   \footnotesize{iNaturalist} & 
                   \footnotesize{OpenImage-O} & 
                   \footnotesize{ImageNet-O} &
                   \footnotesize{Avg.} \\
    \hline
    5.89 & - & 63.05 & 63.64 & 62.86 & 50.54 & 60.02 \\
    \underline{6.39} & - & \textbf{72.29} & 72.2 & 68.18 & 52.28 & 66.23 \\
    - & 0.39 & 61.20 & 75.43 & 75.32 & \textbf{67.73} & 69.92 \\
    - & \underline{0.37} & 61.24 & 76.49 & 75.33 & 67.42 & 69.83 \\
    \underline{6.39} & \underline{0.37} & 69.91 & \textbf{81.06} & \textbf{77.43} & 60.66 & \textbf{72.27} \\
    \bottomrule
    \end{tabular}
    \end{center}}
    \caption{Ablation on early-stop metrics (PSNR and DISTS) for GDM (classifier guidance). We report the best AUROC calculated with DISTS, GMSD~\cite{xue2013gradient} and $\ell_2$ distance for each OODs from the \imagenet benchmark. The best results are in \textbf{bold}. We choose \underline{underlined} thresholds at $t=3/5T$, as stated in Sec.~\ref{sec:classifier-guidance}}
    \label{tab:early-stop}
\end{table}

\begin{figure}[t]
    \centering
    \includegraphics[width=\linewidth]{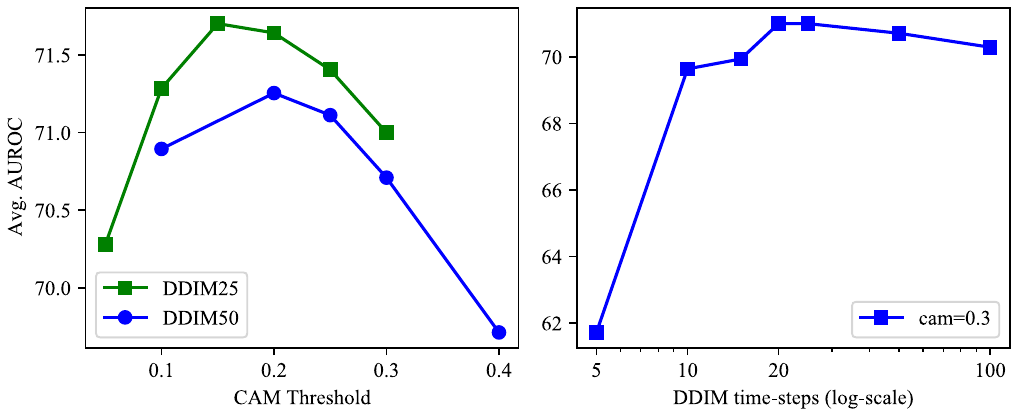}
    \caption{AUROC varies with the CAM cut-point (left) and the number of DDIM steps (right) for LDM (classifier-free guidance)}
    \label{fig:cam-steps}
\end{figure}

\textbf{CAM Cut-point in DSG.}
In Sec.~\ref{sec:classifier-free-guidance}, we propose to use CAM to identify semantic-intensive regions, where the cut-point of the CAM is a hyperparameter.
Typically, the cut-point can be set around 0.2 for various image synthesis settings. To investigate the impact of the CAM cut-point, we use LDM with both DDIM-25 (\ie, DDIM with 25 timesteps) and DDIM-50 for image synthesis and calculate the average AUROC.
As shown in Fig.~\ref{fig:cam-steps} left, the optimal cut-point keeps around 0.2 regardless of the changes of timesteps for image synthesis. A larger cut-point implies a smaller area for conditional generation. Setting a too-small area for conditional generation is insufficient for highlighting semantic mismatch, while applying label guidance globally to all pixels of the image is also unsatisfactory. Empirically, balancing the conditional and unconditional generation at $\operatorname{CAM}\approx 0.2$ achieves the best performance.

\textbf{Different Diffusion Timesteps.}
Since the generation process of diffusion models includes multi-step iterations, the number of timesteps is the key to the trade-off between quality and speed.
For all diffusion-model-based methods including \diffguard, the trade-off still exists even with the DDIM sampler~\cite{song2021denoising}.
To analyze such a trade-off, we test the average AUROC of LDM with different timesteps ranging from 5 to 100.
As shown in Fig.~\ref{fig:cam-steps} right, the AUROC has a non-monotonic correlation with the number of time-steps, and the optimal AUROC is achieved by DDIM-25 empirically.
Although more timesteps generally lead to better synthesis quality, in our case, the timesteps also affect the impact of label guidance.
More guidance steps lead to more significant semantic changes towards the label, potentially leading to more severe distortions. This could explain why fewer steps may perform better for OOD detection.
In addition, it is beneficial to have fewer time-steps for faster inference in practice.
\section{Limitations and Future Work}\label{sec:limiations}
Our method uses a diffusion model for inference, which inherently has a low inference speed due to its iterative nature. Using NVIDIA V100 32GB, GDM (60 steps) and LDM (25 steps) achieve  speeds of 0.05 and 0.53 images/s/GPU respectively. Given that \diffguard relies on diffusion models for both noise addition and denoising, future optimizations should focus on speed improvement in both processes. 

\section{Conclusion}\label{sec:5.conclusion}
In this paper, we investigate the utilization of pre-trained diffusion models for detecting OOD samples through semantic mismatch.
A novel OOD detection framework named \diffguard is proposed, which is compatible with all diffusion models with either classifier guidance or classifier-free guidance.
By guiding the generation process of diffusion models with semantic mismatch, \diffguard accentuates the disparities between InDs and OODs, thus enabling better differentiation.
Moreover, we propose several techniques to enhance different types of diffusion models for OOD detection.
Experimental results show that \diffguard performs well on both \cifar and hard cases from the \imagenet benchmark, without the need for fine-tuning pre-trained diffusion models.

\vspace{0.2cm}
\noindent\textbf{Acknowledgment}. This work was supported in part by the General Research Fund of the Hong Kong Research Grants Council (RGC) under Grant No. 14203521, and in part by the Innovation and Technology Fund under Grant No. MRP/022/20X. We gratefully acknowledge the support of MindSpore, CANN (Compute Architecture for Neural Networks)
and Ascend AI Processor used for this research.

{\small
\bibliographystyle{ieee_fullname}
\bibliography{main}
}

\appendix
\newpage
\appendixpage
\section{Implementation Details of \diffguard}
\textbf{Pre-trained Weights of Diffusion Models.}
On \cifar, we use the same pre-trained model as DiffNB\footnote{https://github.com/luping-liu/DiffOOD}, which is a conditional DDPM~\cite{ho2020denoising} with classifier-free guidance. On \imagenet, we use the unconditional Guided Diffusion Model (GDM)~\cite{dhariwal2021diffusion} and apply classifier guidance\footnote{https://github.com/openai/guided-diffusion}.
While for Latent Diffusion Model (LDM)~\cite{Rombach_2022_CVPR}, we use the model pre-trained on \imagenet\footnote{https://github.com/CompVis/latent-diffusion} with classifier-free guidance.

\textbf{Similarity Metric.}
To measure the similarity of image synthesis and the input, we adopt several similarity metrics~\cite{kastryulin2022piq}, together with the logits from the classifier-under-protection as a commonly considered measure for out-of-distribution (OOD) detection~\cite{liu2023outofdistribution}. Table~\ref{tab:metrics} shows the metrics we consider for different benchmarks.

In general, we find that DISTS performs well on \imagenet. Compared to other low-level metrics (e.g., \ltwo), DISTS provides more robust image-space comparisons. For instance, if the generated image displays different brightness levels from the input, DISTS can offer a more consistent comparison than \ltwo. This is also evidenced by LDM, where DISTS consistently outperforms \ltwo. By contrast, since many similarity metrics (e.g., DISTS, LPIPS) apply pre-trained weights on \imagenet as the feature extractor, they may not be suitable for \cifar directly. Thus, the logits distance works best on \cifar. This result also enables a direct comparison between our \diffguard and DiffNB~\cite{liu2023outofdistribution} (in Table~1), where logits are also utilized as the distance metric.

It is important to note that in the main paper, we report the result only with one generic metric on different benchmarks, without combining different similarity metrics.
In practice, it is feasible to combine multiple metrics for judgment. Such a combination can be either the one employed in Sec.~4.2 and Sec.~4.3, where distinct metrics are treated as additional baselines; or the one presented in \cite{yang2022out}, where various metrics are taken into account, and the rejection of OOD is based on any of them (\ie, work in a tandem manner for OOD rejection).

\textbf{DDIM timesteps.}
In Table~\ref{tab:metrics}, we present the DDIM timesteps utilized in Table~1 and Table~2 of the main paper. Specifically, for \cifar, we opt for the same settings as DiffNB~\cite{liu2023outofdistribution}, using DDIM-50. According to Sec.~4.4, LDM is preferable for fewer DDIM timesteps, resulting in faster inference. In comparison, GDM typically performs better with more DDIM timesteps. To balance the speed and OOD detection performance, we adopt DDIM-100 in the main paper.

\begin{table}[t]
    \small{\begin{center}
    \begin{tabular}{cccc}
    \toprule
    benchmark & model & metrics & DDIM steps \\
    \midrule
    \cifar & DDIM & logits & 50 \\
    \imagenet & GDM & DISTS & 100 \\
    \imagenet & LDM & DISTS & 25 \\
    \bottomrule
    \end{tabular}
    \end{center}}
    \caption{Detailed settings of \diffguard in the main paper for the different benchmarks, including similarity metrics and DDIM timesteps.}
    \label{tab:metrics}
    \end{table}

\begin{table}[ht]
\small{\begin{center}
\begin{tabular}{c|c|cc}
\toprule
method & OOD dataset & \multicolumn{1}{c}{AUROC $\uparrow$} & \multicolumn{1}{c}{FPR@95 $\downarrow$} \\
\midrule
\multirow{5}{*}{GDM(oracle)} & Species & 87.35 & 54.97 \\
 & iNaturalist & 94.15 & 31.60 \\
 & OpenImage-O & 90.97 & 45.94 \\
 & ImageNet-O & 86.22 & 62.20 \\
 \cline{2-4}
 & average & 89.67 & 48.67 \\
 \hline
 \multirow{5}{*}{LDM(oracle)} & Species & 97.38 & 14.41 \\
 & iNaturalist & 97.76 & 12.71 \\
 & OpenImage-O & 95.12 & 25.12 \\
 & ImageNet-O & 95.97 & 22.60 \\
 \cline{2-4}
 & average & 96.56 & 18.71 \\
 \bottomrule
\end{tabular}
\end{center}}
\caption{Results for applying the oracle classifier with \diffguard on the \imagenet benchmark.}
\label{tab:oracle}
\end{table}

\section{Use of the Oracle Classifier on \imagenet}
In Sec.~4.2, we presented the performance of \diffguard on the \cifar benchmark using an oracle classifier. In this section, we demonstrate how \diffguard performs on the \imagenet benchmark with the help of an oracle classifier, as shown in Table~\ref{tab:oracle}. We utilized the same settings as in Table 2. Our results indicate that the performance of LDM and GDM can be significantly improved with an oracle classifier. Since the oracle classifier only provides the predicted label, while GDM relies on the gradient from the classifier, we resort to classifier-under-protection for gradient (\ie, ResNet50). Therefore, its performance may be limited by the incorrect gradient estimation from the classifier. On the other hand, LDM employs classifier-free guidance, and therefore, both AUROC and FPR@95 demonstrate a significant improvement.

\section{More Qualitative Results}
\begin{figure*}[ht]
    \centering
    \includegraphics[width=0.9\linewidth]{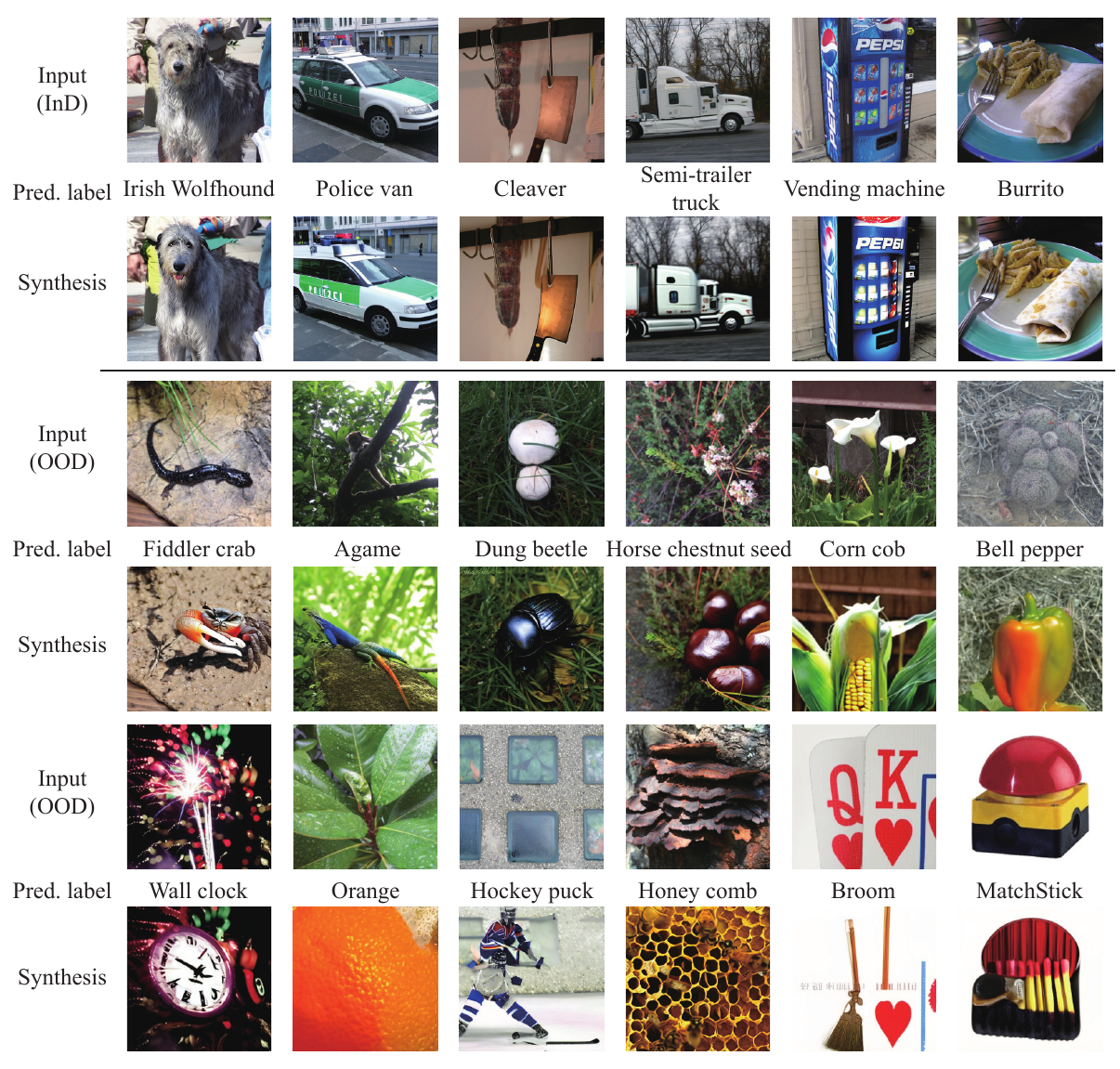}
    \caption{Visualization for InD and OOD cases
    with their syntheses according to the predicted labels. Images are from the \imagenet benchmark. We use LDM in this figure, \ie classifier-free guided diffusion. We can identify a clear similarity difference between InDs and OODs by comparing the inputs with their syntheses.}
    \label{fig:ldm-visual}
\end{figure*}

\begin{figure*}[ht]
    \centering
    \includegraphics[width=0.9\linewidth]{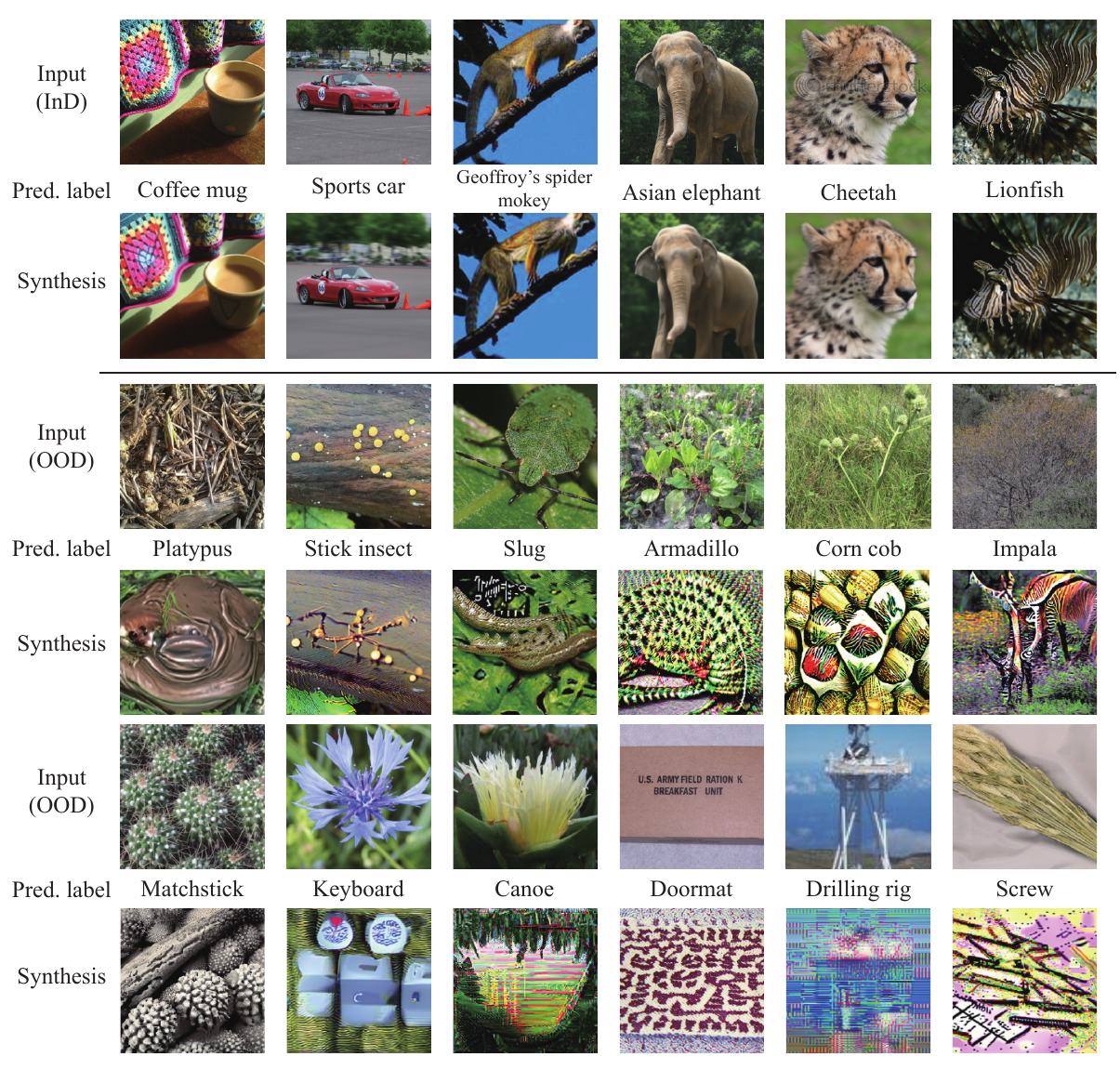}
    \caption{Visualization for InD and OOD cases
    with their syntheses according to the predicted labels. Images are from the \imagenet benchmark. We use GDM in this figure, \ie classifier-guided diffusion. We can identify a clear similarity difference between InDs and OODs by comparing the inputs with their syntheses.}
    \label{fig:gdm-visual}
\end{figure*}

Fig.~\ref{fig:ldm-visual} and Fig.~\ref{fig:gdm-visual} display the image syntheses by \diffguard with LDM and GDM, respectively. The visualization reveals that \diffguard can effectively produce analogous images in InD scenarios, while emphasizing the semantic mismatch in OOD scenarios. Regarding the comparison between GDM and LDM, we notice that GDM occasionally incorporates unrealistic features from the classifier-under-protection in the synthesized images, while LDM consistently generates photo-realistic syntheses, even in OOD cases. Such a phenomenon on GDM motivates us to employ adaptive early-stop (AES) in Sec.~3.2.1, Tech \#2. Despite that LDM sometimes alters InD samples, GDM does not. This further justifies \diffguard to extract and use the information from the classifier-under-protection for LDM, as stated in Sec.~3.2.2, Tech \#3. As shown in Fig.~\ref{fig:ldm-visual}, only certain details are modified after applying \diffguard with Tech \#3, while the main structure and content are preserved.

\section{Failure Case Analysis}
\begin{figure}[t]
    \centering
    \includegraphics[width=0.95\linewidth]{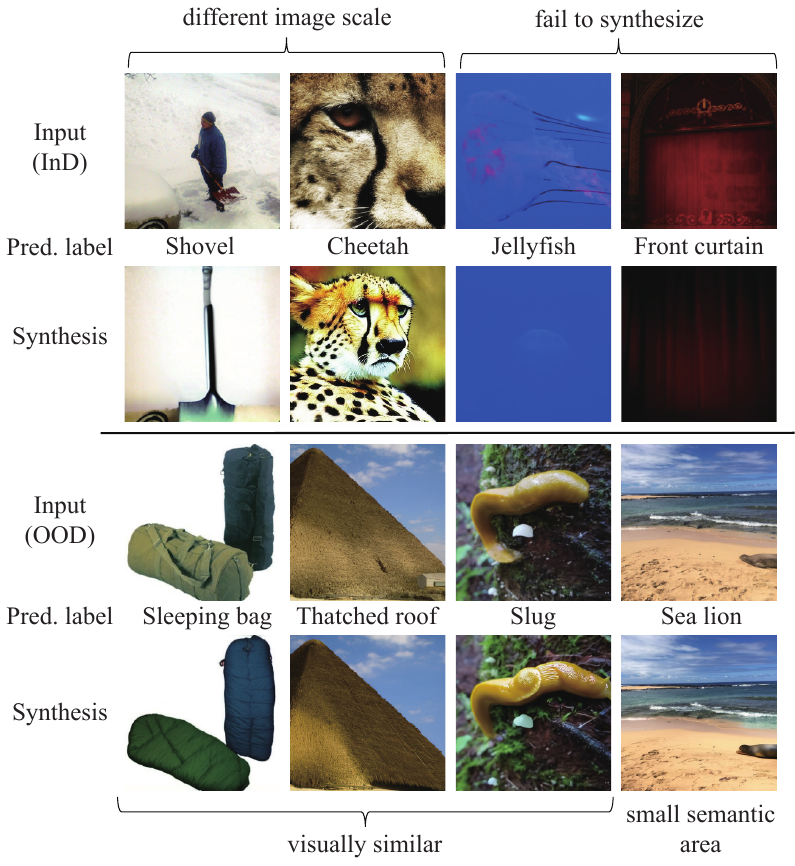}
    \caption{Some failure cases for \diffguard. \diffguard may fail when image synthesis fails for InDs or when the content of OODs is indeed visually similar to InDs' semantics.}
    \label{fig:failure}
\end{figure}

We present some failure cases of \diffguard in Fig.~\ref{fig:failure}. For InDs, these failures are mainly due to image synthesis problems. For example, we observe some test cases exhibit different fields of view from common cases, which makes \diffguard difficult to maintain their original content. Additionally, certain classes (e.g. jellyfish and front curtain) tend to be monochrome or dark, which could cause generative models to fail in synthesizing such images. To address these issues, better generative models may help.

Regarding OODs, the major problem is that some cases appear visually similar to InD classes, or the area of semantic mismatch is limited. It is worth noting that \diffguard can successfully depict the target semantics in these cases, but the image synthesis still looks similar to the input, resulting in difficulty to detect them by similarity measurements. To solve such cases, one possible solution is to utilize better similarity metrics for detailed comparisons (e.g. feature distance from a model trained with contrastive learning~\cite{wang2020hypersphere}).

\end{document}